\documentclass[11pt]{article}

\usepackage[final]{acl}

\usepackage{times}
\usepackage{latexsym}

\usepackage[T1]{fontenc}

\usepackage[utf8]{inputenc}

\usepackage{microtype}

\usepackage{inconsolata}

\usepackage{graphicx}
\usepackage{amsmath}
\usepackage{amssymb}
\usepackage[normalem]{ulem}
\usepackage{multirow}
\usepackage{makecell}
\usepackage[table]{xcolor}
\definecolor{CustomGreen}{RGB}{198, 239, 206}
\definecolor{CustomOrange}{RGB}{248, 203, 173}

\usepackage{booktabs}
\usepackage{amsmath}
\usepackage{array}
\usepackage{arydshln}
\usepackage{algorithm}
\usepackage{algpseudocode}
\usepackage[most]{tcolorbox}

\usepackage{mathtools}
\title{Analyzing and Mitigating Cross-Lingual Degradation in Multilingual Medical VQA}

\author{
  \textbf{Jingbo Wang},
  \textbf{Sendong Zhao}\thanks{Corresponding author.},
  \textbf{Haochun Wang},
  \textbf{Bing Qin},
  \textbf{Ting Liu}
  \\
  Research Center for Social Computing and Interactive Robotics \\
  Harbin Institute of Technology, China \\
  \texttt{\{jingbowang,sdzhao\}@ir.hit.edu.cn}
}

\begin{document}
\maketitle
\begin{abstract}
Medical visual question answering (VQA) is a crucial task in clinical AI, yet its evaluation has so far centered almost exclusively on English, limiting its relevance to linguistically diverse patients and clinicians. 
Recent multilingual medical VQA benchmarks show that large vision-language models (LVLMs) degrade in non-English languages, but lack a fine-grained analysis of how cross-lingual variation affects the distinct capabilities that medical VQA requires. 
To this end, we construct a multilingual medical VQA benchmark over eight languages, organized into four representative scenarios that isolate the core capabilities medical VQA requires. 
Evaluating five open- and closed-source LVLMs, we find that cross-lingual degradation is not uniform but highly scenario-dependent.
We therefore propose MedVL-XLRepE, a training-free scenario-aware representation engineering method, leveraging LVLMs' superior English medical VQA capability to steer non-English representations toward their English counterparts at inference time. 
Across three LVLMs and eight languages, MedVL-XLRepE consistently mitigates cross-lingual degradation, with gains of up to 6.33\%.

\end{abstract}

\section{Introduction}

Medical visual question answering (VQA), which aims to interpret medical images and answer clinical queries, is a crucial research area in clinical AI~\citep{lau2018dataset,he2020pathvqa,li2023llava}. Driven by the rapid progress of large vision-language models (LVLMs)~\citep{chen2024towards,deepseekai2026deepseekv4}, recent systems have achieved substantial progress on medical VQA~\citep{hu2024omnimedvqa,dong2025generative}. Despite these remarkable advances, the medical evaluation of LVLMs has so far centered almost exclusively on English, which limits their applicability to a linguistically diverse population of patients and clinicians.

To evaluate whether LVLMs can serve multilingual populations, recent studies have constructed multilingual medical VQA benchmarks. These benchmarks draw medical images and questions from the national medical examinations and real-world clinical consultations of several countries, and evaluate the cross-lingual performance of LVLMs~\citep{WorldMedQA-V2024,riccio2025multilingual,yim2024dermavqa}. Although these benchmarks show that the medical VQA performance of LVLMs degrades in non-English languages, their analysis of this degradation has two key limitations. First, they measure only the \emph{overall success rate} on medical VQA, lacking fine-grained analysis of how cross-lingual variation affects the distinct capabilities that medical VQA requires. Second, their \emph{language coverage is limited}, lacking analysis of how this degradation generalizes across a broader range of languages.

To address these limitations, we construct a multilingual medical VQA benchmark covering eight languages. Guided by the core capabilities that medical VQA requires~\citep{lin2023medical,pahud2024orchestrating,acosta2022multimodal}, we organize the benchmark into four representative scenarios: Perceptual Recognition, Attribute-Aware Recognition, Sequential Images Understanding, and Vision-Text Integrated Reasoning, each isolating one such capability. 
Evaluations across five open- and closed-source LVLMs reveal that cross-lingual degradation is not uniform but highly scenario-dependent, indicating that language affects different medical reasoning capabilities unevenly. 
Such scenario-dependent cross-lingual degradation cannot be addressed by a single uniform correction, but requires a scenario-specific method instead.

To leverage the superior medical VQA capabilities of LVLMs in English for improving non-English performance, we propose MedVL-XLRepE, a training-free, scenario-aware cross-lingual representation engineering method. For each scenario, MedVL-XLRepE leverages a language vector and scenario-specific medical vectors from the representation gap between parallel English and target-language inputs, steering non-English representations toward their English counterparts at inference time. Experiments across three LVLMs and eight languages show that MedVL-XLRepE achieves consistent improvements, with gains of up to 6.33\%, demonstrating its effectiveness in mitigating cross-lingual degradation in medical VQA.

Our contributions are summarized as follows:
\begin{itemize}
    \item We construct a multilingual medical VQA benchmark covering eight languages and organized into four scenarios that isolate the core capabilities medical VQA requires.
    \item We present a comprehensive analysis of multilingual medical VQA across open- and closed-source LVLMs, revealing that cross-lingual degradation is strongly scenario-dependent.
    \item We propose MedVL-XLRepE, a scenario-aware representation engineering method that effectively mitigates cross-lingual degradation in medical VQA.
\end{itemize}

\section{Related Work}

\subsection{Multilingual Medical VQA Benchmarks}

Medical VQA has become a core task for evaluating clinical AI~\citep{lau2018dataset,he2020pathvqa,li2023llava,hu2024omnimedvqa}. Existing benchmarks, however, remain predominantly English-centric. Recent studies show that models excelling at English degrade substantially in other languages~\citep{zhang2023m3exam,schmidt2025mvl}, underscoring the need to evaluate medical VQA beyond English.
To this end, a few multilingual medical VQA benchmarks have recently emerged. WorldMedQA-V~\citep{WorldMedQA-V2024} and MMMED~\citep{riccio2025multilingual} draw image-based questions from the national medical examinations of several countries, whereas DermaVQA~\citep{yim2024dermavqa} and MEDIQA-M3G~\citep{yim2024overview} target real-world dermatology consultations.
Nevertheless, these benchmarks remain too coarse to reveal how cross-lingual variation affects the clinical capabilities that medical VQA requires.

\subsection{Cross-lingual Representation Engineering}

Prior works have demonstrated that semantically aligned inputs across languages occupy divergent regions in the latent space, a divergence associated with cross-lingual performance gaps~\citep{chang2022geometry,zhao2024large,peng2025debiasing}. Complementing this, studies in representation engineering have established that hidden-state interventions offer an effective means of steering model outputs at inference time~\citep{zou2023representation,turner2023steering,li2023inference}. Motivated by these observations, recent work derives a steering vector that pulls non-English representations toward their English counterparts at inference, thereby narrowing the cross-lingual gap~\citep{li2025unlocking,chen2025mpr}.
However, these methods rely on a single type of steering vector applied uniformly across all inputs, and thus cannot match cross-lingual degradation that differs across clinical capabilities.

\section{Multilingual Medical VQA Benchmark}
\label{sec:analysis}

\begin{figure}[t]
\centering
\includegraphics[width=\linewidth]{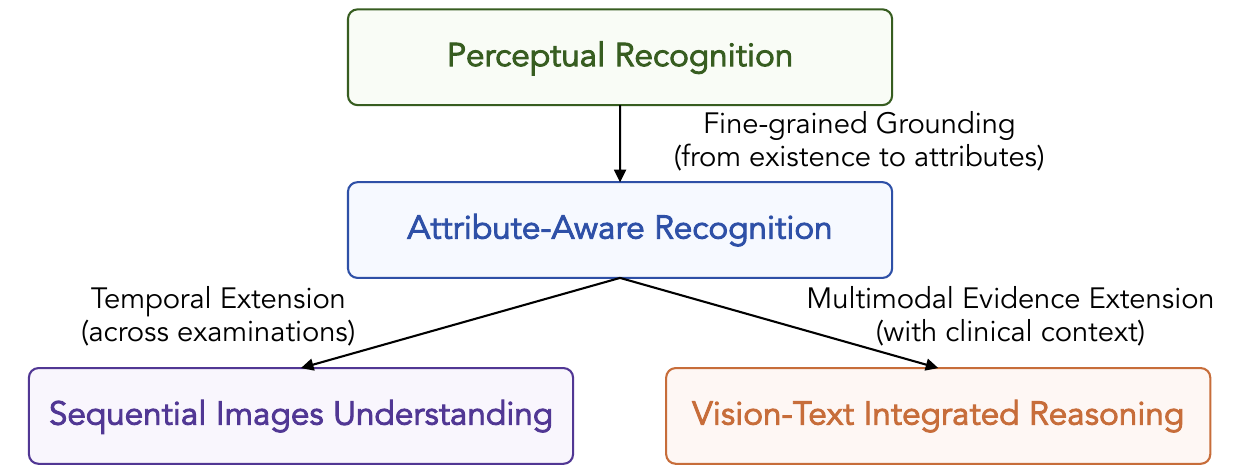}
\caption{Medical VQA scenarios studied in this work.}
\label{fig:scenario}
\end{figure}

In this section, we study cross-lingual medical VQA in a controlled setting where visual evidence and clinical intent are comparable across languages. We first organize the task into representative clinical scenarios and construct a semantically aligned multilingual benchmark, then analyze performance variation across models, languages, and scenarios. The goal is to identify which forms of medical multimodal reasoning are most sensitive to language change, and to use these findings to motivate the method in the next section.
\begin{figure*}[htb]
    \centering
    \includegraphics[width=0.9\textwidth]{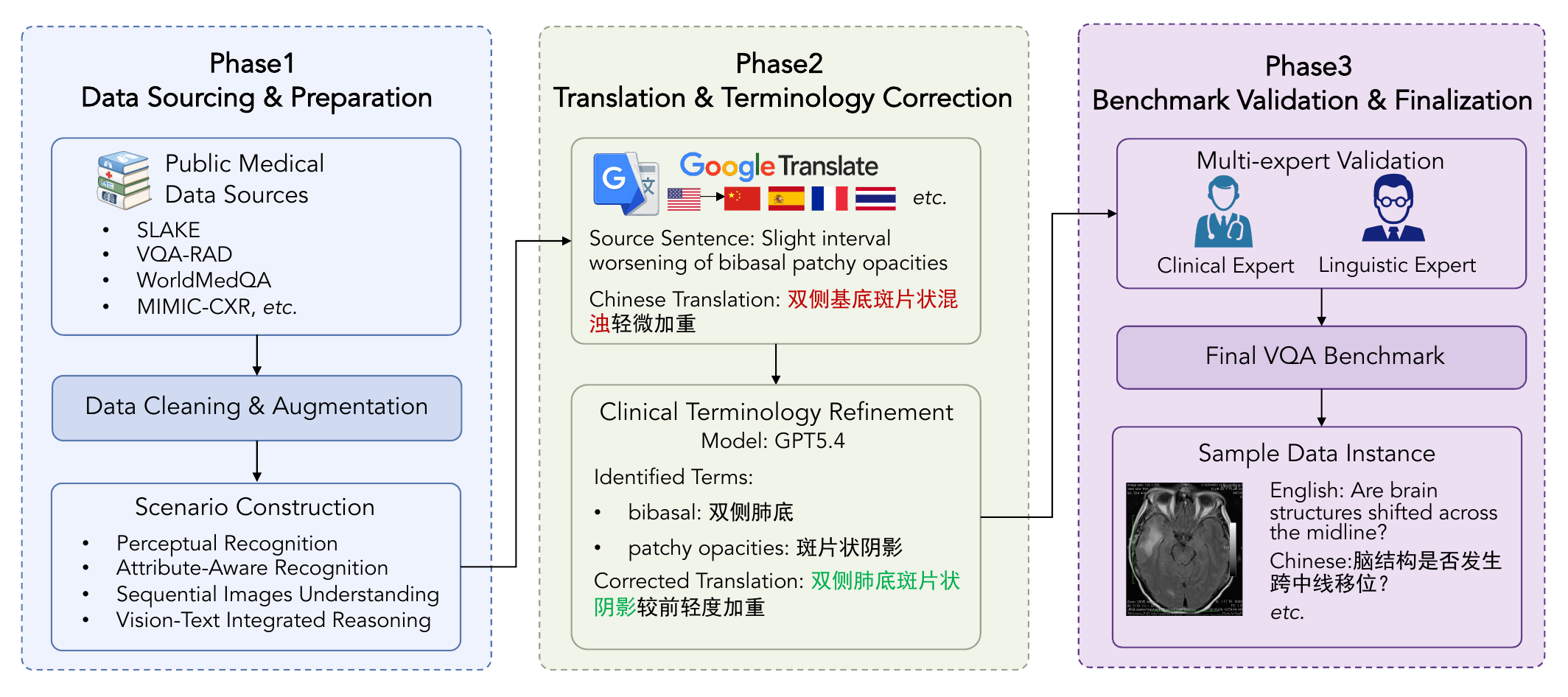}
    \caption{Construction pipeline of the multilingual medical VQA benchmark: data sourcing and scenario-wise reorganization, multilingual translation and terminology alignment, and benchmark validation and finalization.}
    \label{fig:data_construction}
\end{figure*}

\subsection{Medical VQA Problem Definition}
Medical VQA requires the model to answer clinically meaningful questions from medical images, often by identifying relevant findings and their attributes \citep{lin2023medical}, comparing current examinations with prior studies \citep{pahud2024orchestrating}, and interpreting imaging evidence together with accompanying clinical text \citep{acosta2022multimodal}. In practice, these capability demands recur in four clinical scenarios centered on finding recognition, attribute characterization, temporal comparison, and image-text interpretation. As illustrated in Figure~\ref{fig:scenario}, we therefore define four scenarios for medical VQA, referred to as Perceptual Recognition, Attribute-Aware Recognition, Sequential Images Understanding, and Vision-Text Integrated Reasoning.

\paragraph{Perceptual Recognition (PR)} PR refers to single-image questions that ask whether a visible finding, structure, or abnormal pattern is present. It reflects the most basic perceptual requirement in medical VQA, since clinically meaningful answering often begins with identifying the relevant visual finding.

\paragraph{Attribute-Aware Recognition (AAR)} AAR extends basic finding recognition by requiring the model to determine clinically relevant attributes of a finding. Such attributes include laterality, anatomical location, size, extent, severity, morphology, and spatial relation, all of which are essential to medically precise description and interpretation. The challenge is not only to identify the finding itself but also to relate the queried attribute to the visual finding.

\paragraph{Sequential Images Understanding (SIU)} SIU concerns questions that require the model to compare temporally ordered studies and determine how a finding evolves over time. This scenario is needed when clinical interpretation depends not only on the current image, but also on change relative to prior examinations, where progression, improvement, or stability can be diagnostically decisive. The central challenge in SIU is to align corresponding medical content across time points and determine the relevant change between studies, rather than to interpret each image in isolation.

\paragraph{Vision-Text Integrated Reasoning (VTI)} VTI extends medical VQA beyond image-only questions to cases where the answer depends on both the image and complementary clinical text, such as patient history or medical records. It reflects clinical questions for which visual content alone is insufficient and textual medical context is necessary for interpretation. This scenario places a greater demand on accurate understanding of medical text in addition to the image.

\subsection{Benchmark Construction}
\label{sec:benchmark_construction}
To analyze cross-lingual effects, we construct a multilingual medical VQA benchmark in which visual evidence and clinical intent are controlled across languages, making language the primary variable under comparison. The construction goal is not merely to aggregate existing medical datasets, but to reorganize them into scenario-aligned probes that localize where language variation perturbs multimodal medical reasoning. As illustrated in Figure~\ref{fig:data_construction}, we organize benchmark construction into three stages to preserve cross-lingual comparability while maintaining medical fidelity.

\newcolumntype{C}[1]{>{\centering\arraybackslash}p{#1}}
\newcolumntype{L}[1]{>{\raggedright\arraybackslash}p{#1}}
\begin{table*}[!ht]
\centering
\scriptsize
\renewcommand{\arraystretch}{0.95}
\setlength{\tabcolsep}{4pt}
\resizebox{\textwidth}{!}{%
\begin{tabular}{C{2.6cm} C{0.9cm} C{0.8cm} C{0.8cm} C{0.8cm} C{0.8cm} C{0.8cm} C{0.8cm} C{0.8cm} C{0.8cm}}
\toprule
\textbf{Model} & \textbf{Scenario} & \textbf{EN} & \textbf{ZH} & \textbf{ES} & \textbf{FR} & \textbf{JA} & \textbf{TH} & \textbf{AR} & \textbf{BG} \\
\midrule
\rowcolor[gray]{0.95} \multicolumn{10}{c}{\textit{Open-source LVLMs}} \\
\multirow{4}{=}{Gemma3-12B-IT} & PR & 66.67 & \textbf{67.49} & 64.21 & 64.48 & 62.57 & 64.21 & 65.85 & \underline{59.84} \\
& AAR & \textbf{52.54} & 50.85 & 44.07 & 39.83 & 46.61 & 47.46 & 46.61 & \underline{38.98} \\
& SIU & \textbf{43.66} & 41.79 & 40.30 & 42.54 & 40.30 & 40.67 & \underline{39.55} & 40.30 \\
& VTI & \textbf{54.23} & 46.48 & 48.24 & 47.18 & 44.01 & 45.42 & \underline{40.14} & 46.83 \\
\midrule
\multirow{4}{=}{Qwen3.5-9B} & PR & \textbf{75.96} & 72.95 & \textbf{75.96} & 74.86 & \underline{71.31} & 75.41 & 73.50 & 72.68 \\
& AAR & \textbf{73.73} & 66.10 & 69.49 & 70.34 & 69.49 & \underline{65.25} & 67.80 & 68.64 \\
& SIU & \textbf{66.42} & 61.19 & 61.57 & \underline{57.09} & 61.19 & 61.57 & 58.21 & 60.45 \\
& VTI & \textbf{64.79} & 61.27 & 62.68 & 63.03 & \underline{56.34} & 57.39 & 58.45 & 62.68 \\
\midrule
\multirow{4}{=}{InternVL3.5-14B-Instruct} & PR & \textbf{71.31} & 65.57 & 66.67 & 65.85 & 58.74 & 62.57 & \underline{56.01} & 66.39 \\
& AAR & \textbf{66.10} & 59.32 & 61.02 & \underline{52.54} & 61.86 & 53.39 & 56.78 & 61.86 \\
& SIU & \textbf{54.85} & 49.63 & 53.36 & 53.73 & 47.01 & 51.49 & \underline{45.90} & 50.37 \\
& VTI & 56.69 & \textbf{57.75} & 53.52 & 50.35 & 50.00 & 42.96 & \underline{40.14} & 47.54 \\
\midrule
\rowcolor[gray]{0.95} \multicolumn{10}{c}{\textit{Closed-source LVLMs}} \\
\multirow{4}{=}{GPT-5.4-mini} & PR & \textbf{78.69} & 74.86 & 76.23 & 75.14 & \underline{72.68} & 77.32 & 73.50 & 75.96 \\
& AAR & 75.42 & 74.58 & 72.03 & \underline{69.49} & 74.58 & 72.03 & \textbf{77.12} & 73.73 \\
& SIU & 56.72 & 52.24 & \textbf{57.46} & 52.61 & 53.73 & 55.60 & 55.60 & \underline{50.75} \\
& VTI & \textbf{75.70} & 75.35 & 74.65 & 73.59 & \underline{72.89} & 73.24 & 74.65 & \underline{72.89} \\
\midrule
\multirow{4}{=}{Gemini-3-Flash-Preview} & PR & \textbf{79.51} & \underline{77.05} & 78.69 & 77.32 & 78.42 & \textbf{79.51} & 77.32 & 77.60 \\
& AAR & 77.97 & 76.27 & \textbf{81.36} & \underline{72.88} & 77.12 & 79.66 & 79.66 & 78.81 \\
& SIU & \textbf{74.63} & 70.15 & 73.88 & 72.39 & \underline{69.40} & 70.90 & 71.27 & 71.64 \\
& VTI & \textbf{89.08} & 86.27 & 88.38 & 86.27 & 85.92 & 86.62 & 86.62 & \underline{84.15} \\
\bottomrule
\end{tabular}%
}
\caption{Performance (\%) of open- and closed-source LVLMs across four medical VQA scenarios and eight languages in our multilingual VQA benchmark. \textbf{Bold} and \underline{underlined} values indicate the best and worst performance.}
\small
\label{tab:performance_result} 
\end{table*}

\paragraph{Step 1: Data Sourcing and Scenario-wise Reorganization} We begin with public medical multimodal resources, including VQA-RAD \citep{lau2018dataset}, WorldMedQA-V \citep{WorldMedQA-V2024}, MMXU \citep{mu2025mmxu}, and MIMIC-CXR \citep{johnson2019mimic}, and retain only those samples for which the image evidence, clinical intent, and answer supervision remain comparable after multilingual translation. The retained samples are then reorganized by task mechanism rather than by original dataset identity, and the final benchmark is formulated using closed-form questions only, including binary yes/no and multiple-choice formats, to reduce answer-form variability across languages.

\paragraph{Step 2: Multilingual Translation and Terminology Alignment} For the resulting English-source samples, we construct parallel multilingual questions spanning both mid-to-high-resource languages (Chinese, Spanish, French, and Japanese) and comparatively lower-resource languages (Thai, Arabic, and Bulgarian), while keeping the same underlying visual evidence and answer supervision across languages. We first use Google Translate to obtain draft translations at scale. However, as illustrated in Figure~\ref{fig:data_construction}, these drafts do not always preserve medical terminology with sufficient precision. GPT-5.4 \citep{gpt54} is then used for terminology-focused refinement, with the goal of improving medical terminology accuracy while preserving the clinical intent of the original question.

\paragraph{Step 3: Benchmark Validation and Finalization} After translation and terminology refinement, the multilingual items are reviewed by medical and linguistic experts before inclusion in the final benchmark. The review verifies that each item preserves the intended medical meaning across languages, remains grounded in the image evidence required for the question, and maintains a stable answer mapping after multilingual construction. The detailed scenario-wise data statistics are provided in the appendix~\ref{sec:data_stats}. 

\subsection{Experiment Setup}

\paragraph{Baseline}
We select baselines from two model types: (1) \textbf{Open-source LVLMs}: Gemma3-12B-IT \citep{Gemma3}, Qwen3.5-9B \citep{qwen3.5}, and InternVL3.5-14B-Instruct \citep{wang2025internvl3}; (2) \textbf{Closed-source LVLMs}: GPT-5.4-mini \citep{gpt54} and Gemini-3-Flash-Preview \citep{gemini3flash}.

\paragraph{Implementation Details}
For open-source models, evaluation is conducted on a server with 8 $\times$ NVIDIA H20 GPUs. For stable reproduction, we set the decoding temperature to 0 in all experiments.

\subsection{Cross-lingual Evaluation Analysis}
\label{sec:eval_analysis}

\paragraph{Overall Multilingual Gaps Are Clear, and Closed-Source Models Are More Stable}
As shown in Table~\ref{tab:performance_result}, multilingual performance gaps are clearly observable in the overall results. Across languages, the overall ranking is EN, ES, ZH, TH, BG, FR, JA, and AR. In particular, English attains the highest mean accuracy at 67.73\%, whereas Arabic yields the lowest at 62.23\%, corresponding to a gap of 5.50\%.
Thus, even when visual evidence and question intent are controlled across languages, language changes still lead to measurable performance differences in medical VQA.
As shown in Table~\ref{tab:source_group_aggregated_language_performance}, closed-source models consistently outperform open-source models, achieving an average of 74.03\% compared with 57.22\% for open-source models. The same table further shows that closed-source models exhibit more stable cross-lingual performance: relative to their English average of 75.97\%, the largest gap to a non-English language is 3.51\%, whereas the corresponding English average for open-source models is 62.25\%, with the largest gap reaching 8.17\%.

\paragraph{Cross-Lingual Degradation Is Not Uniform Across Scenarios}
As shown in Table~\ref{tab:scenario_aggregated_language_performance}, cross-lingual degradation varies substantially across scenarios rather than appearing as a uniform reduction throughout the benchmark. PR shows a relatively small cross-lingual span of 5.68\%. When the task shifts from recognizing whether a finding is present to determining medically relevant attributes of that finding, the span increases to 8.14\% in AAR, indicating that language variation more strongly affects attribute grounding than basic finding recognition. When the task further requires integrating visual evidence with complementary clinical text, VTI still exhibits a large span of 8.10\%, suggesting that cross-lingual differences remain pronounced in image-text evidence alignment. SIU, in contrast, yields a smaller span of 5.15\%, indicating that introducing temporal comparison across multiple images does not by itself lead to the largest cross-lingual gap.

\paragraph{The Hardest Scenario Is Not the Most Language-Sensitive}
As shown in Table~\ref{tab:scenario_aggregated_language_performance}, multilingual fragility cannot be reduced to intrinsic task difficulty. SIU is the hardest scenario in the benchmark, with the lowest average accuracy at 55.80\%, yet it is also the most stable across languages, with a span of only 5.15\%. AAR and VTI show a different profile: both are easier than SIU in overall accuracy, at 65.08\% and 63.61\%, but they exhibit substantially larger cross-lingual spans of 8.14\% and 8.10\%, respectively. This direct contrast shows that the strongest multilingual instability does not arise in the intrinsically hardest scenario, but in scenarios whose required medical reasoning is more vulnerable to language change.

\paragraph{Language Resource Level Does Not Determine Scenario-Level Performance}
Language resource level does not map directly onto multilingual medical VQA performance. Although English is the strongest language overall at 67.73\%, the distinction between mid-to-high-resource languages (Chinese, Spanish, French, and Japanese) and comparatively lower-resource languages (Thai, Arabic, and Bulgarian) is much less pronounced, with only a 1.01\% difference in aggregate (63.83\% vs.\ 62.82\%). Moreover, relative performance can reverse across resource groups once scenario is taken into account: Japanese is stronger than Thai on AAR (65.93\% vs.\ 63.56\%), whereas Thai outperforms Japanese on PR (71.80\% vs.\ 68.74\%). The full scenario-wise language rankings are provided in Table~\ref{tab:language_rankings_by_scenario}.

\section{MedVL-XLRepE}

Section~\ref{sec:eval_analysis} shows that cross-lingual performance gaps in medical VQA are scenario-dependent rather than uniform. This suggests that language variation should be handled with selective alignment rather than a single global adjustment. Recent work on representation engineering has shown that lightweight inference-time interventions on internal activations can steer model behavior without updating model parameters \citep{turner2023steering,zou2023representation}; more recent multilingual studies further suggest that similar interventions can help narrow cross-lingual gaps in perception and reasoning \citep{li2025unlocking,chen2025mpr}. Motivated by this, we propose MedVL-XLRepE, a scenario-aware Medical Vision-Language Cross-lingual Representation Engineering method for multilingual medical VQA.

\begin{figure*}[t]
    \centering
    \includegraphics[width=0.9\textwidth]{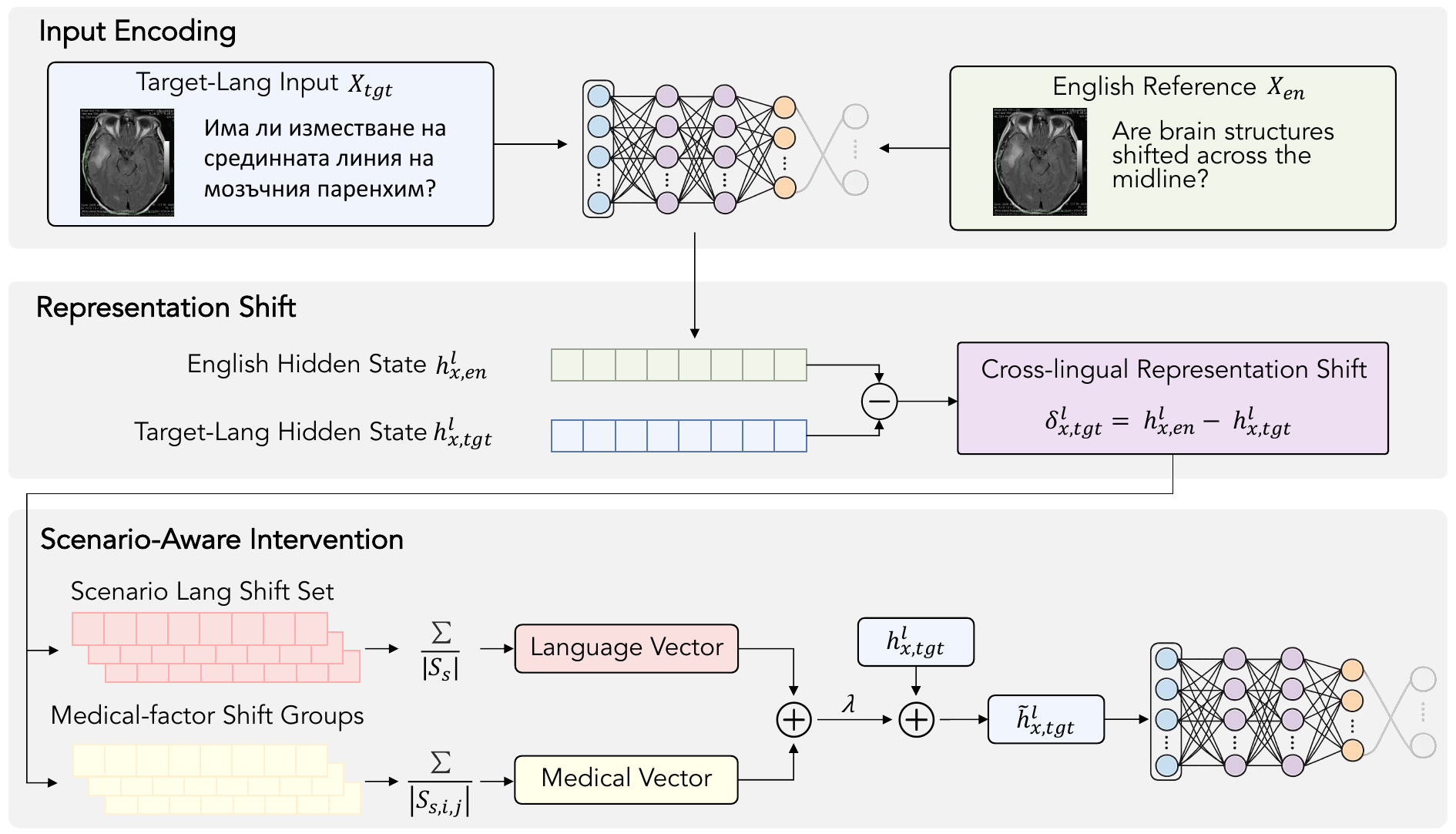}
    \caption{Overview of MedVL-XLRepE. The hidden-state difference between paired English and target-language inputs defines a cross-lingual shift $\delta_{x,tgt}^{(l)}$. Aggregating $\delta_{x,tgt}^{(l)}$ over the scenario set yields a language vector and a medical vector, whose sum is added to $h_{x,tgt}^{(l)}$ to obtain the intervened representation $\tilde{h}_{x,tgt}^{(l)}$.}
    \label{fig:medvl_xlrepe}
\end{figure*}

\subsection{Preliminaries}

For each sample $x$, let $I$ denote the medical image, and let $T_{en}$ and $T_{tgt}$ denote the English and target-language textual inputs. The corresponding multimodal inputs are defined as
\begin{equation}
X_{en} = (I, T_{en}), \qquad X_{tgt} = (I, T_{tgt})
\end{equation}
Due to the autoregressive nature of LVLMs, the final input token's hidden state provides a compact summary of the encoded multimodal context. We therefore use this token as the anchor representation for cross-lingual comparison. Specifically, let $h_{x,en}^{(l)}$ and $h_{x,tgt}^{(l)}$ denote the hidden states of the final input token for sample $x$ at layer $l$ under $X_{en}$ and $X_{tgt}$, respectively. We then define the cross-lingual representation shift as
\begin{equation}
\delta_{x,tgt}^{(l)} = h_{x,en}^{(l)} - h_{x,tgt}^{(l)}
\end{equation}
This difference captures the sample-level English-target representation shift at layer $l$ and serves as the basic quantity for the vector construction below.

\subsection{Scenario-Aware Intervention Vector}
\label{sec:vector_construction}

The cross-lingual shift within each scenario stems from two distinct sources: a general language-level shift between English and the target language, and finer shifts shaped by the scenario's clinical factors.

\paragraph{Language-Level Vector}
For each target language and scenario $s$, we average the sample-level representation differences over all samples in that scenario:
\begin{equation}
v_{lang,s}^{(l)}
=
\frac{1}{|S_{s}|}
\sum_{x\in S_{s}}
\delta_{x,tgt}^{(l)}
\end{equation}
where $S_{s}$ denotes the set of samples belonging to scenario $s$. This vector captures the systematic English-target shift common to all samples in scenario $s$ at layer $l$.

\newcommand{\gainbox}[2]{{\setlength{\fboxsep}{1pt}\colorbox{#1}{#2}}}

\begin{table*}[t]
\centering
\scriptsize
\renewcommand{\arraystretch}{0.95}
\setlength{\tabcolsep}{3pt}
\resizebox{\textwidth}{!}{%
\begin{tabular}{>{\centering\arraybackslash}m{2.6cm} >{\centering\arraybackslash}m{0.9cm} >{\centering\arraybackslash}m{1.05cm} >{\raggedright\arraybackslash}m{0.95cm} >{\raggedright\arraybackslash}m{1.45cm} >{\raggedright\arraybackslash}m{1.45cm} >{\raggedright\arraybackslash}m{1.45cm} >{\raggedright\arraybackslash}m{1.45cm} >{\raggedright\arraybackslash}m{1.45cm} >{\raggedright\arraybackslash}m{1.45cm} >{\raggedright\arraybackslash}m{1.45cm}}
\toprule
\textbf{Model} & \textbf{Scenario} & \textbf{XLRepE} & \textbf{EN} & \textbf{ZH} & \textbf{ES} & \textbf{FR} & \textbf{JA} & \textbf{TH} & \textbf{AR} & \textbf{BG} \\
\midrule
\multirow{4}{=}{Gemma3-12B-IT} & PR & \makecell{$\times$ \\ $\checkmark$} & \makecell[l]{66.67 \\ \textemdash} & \makecell[l]{67.49 \\ 68.58 \gainbox{CustomGreen}{$\uparrow$1.09}} & \makecell[l]{64.21 \\ 65.03 \gainbox{CustomGreen}{$\uparrow$0.82}} & \makecell[l]{64.48 \\ 64.48 \gainbox{yellow!30}{$\uparrow$0.00}} & \makecell[l]{62.57 \\ 63.11 \gainbox{CustomGreen}{$\uparrow$0.54}} & \makecell[l]{64.21 \\ 66.94 \gainbox{CustomGreen}{$\uparrow$2.73}} & \makecell[l]{65.85 \\ 68.03 \gainbox{CustomGreen}{$\uparrow$2.18}} & \makecell[l]{59.84 \\ 63.93 \gainbox{CustomGreen}{$\uparrow$4.09}} \\
\cmidrule(lr){2-11}
 & AAR & \makecell{$\times$ \\ $\checkmark$} & \makecell[l]{52.54 \\ \textemdash} & \makecell[l]{50.85 \\ 52.54 \gainbox{CustomGreen}{$\uparrow$1.69}} & \makecell[l]{44.07 \\ 44.92 \gainbox{CustomGreen}{$\uparrow$0.85}} & \makecell[l]{39.83 \\ 44.92 \gainbox{CustomGreen}{$\uparrow$5.09}} & \makecell[l]{46.61 \\ 46.61 \gainbox{yellow!30}{$\uparrow$0.00}} & \makecell[l]{47.46 \\ 50.00 \gainbox{CustomGreen}{$\uparrow$2.54}} & \makecell[l]{46.61 \\ 47.46 \gainbox{CustomGreen}{$\uparrow$0.85}} & \makecell[l]{38.98 \\ 44.07 \gainbox{CustomGreen}{$\uparrow$5.09}} \\
\cmidrule(lr){2-11}
 & SIU & \makecell{$\times$ \\ $\checkmark$} & \makecell[l]{43.66 \\ \textemdash} & \makecell[l]{41.79 \\ 43.66 \gainbox{CustomGreen}{$\uparrow$1.87}} & \makecell[l]{40.30 \\ 43.28 \gainbox{CustomGreen}{$\uparrow$2.98}} & \makecell[l]{42.54 \\ 44.03 \gainbox{CustomGreen}{$\uparrow$1.49}} & \makecell[l]{40.30 \\ 42.54 \gainbox{CustomGreen}{$\uparrow$2.24}} & \makecell[l]{40.67 \\ 44.03 \gainbox{CustomGreen}{$\uparrow$3.36}} & \makecell[l]{39.55 \\ 42.54 \gainbox{CustomGreen}{$\uparrow$2.99}} & \makecell[l]{40.30 \\ 43.28 \gainbox{CustomGreen}{$\uparrow$2.98}} \\
\cmidrule(lr){2-11}
 & VTI & \makecell{$\times$ \\ $\checkmark$} & \makecell[l]{54.23 \\ \textemdash} & \makecell[l]{46.48 \\ 47.54 \gainbox{CustomGreen}{$\uparrow$1.06}} & \makecell[l]{48.24 \\ 51.76 \gainbox{CustomGreen}{$\uparrow$3.52}} & \makecell[l]{47.18 \\ 49.30 \gainbox{CustomGreen}{$\uparrow$2.12}} & \makecell[l]{44.01 \\ 47.54 \gainbox{CustomGreen}{$\uparrow$3.53}} & \makecell[l]{45.42 \\ 46.13 \gainbox{CustomGreen}{$\uparrow$0.71}} & \makecell[l]{40.14 \\ 41.55 \gainbox{CustomGreen}{$\uparrow$1.41}} & \makecell[l]{46.83 \\ 47.18 \gainbox{CustomGreen}{$\uparrow$0.35}} \\
\midrule
\multirow{4}{=}{Qwen3.5-9B} & PR & \makecell{$\times$ \\ $\checkmark$} & \makecell[l]{75.96 \\ \textemdash} & \makecell[l]{72.95 \\ 74.32 \gainbox{CustomGreen}{$\uparrow$1.37}} & \makecell[l]{75.96 \\ 77.60 \gainbox{CustomGreen}{$\uparrow$1.64}} & \makecell[l]{74.86 \\ 76.23 \gainbox{CustomGreen}{$\uparrow$1.37}} & \makecell[l]{71.31 \\ 75.41 \gainbox{CustomGreen}{$\uparrow$4.10}} & \makecell[l]{75.41 \\ 77.32 \gainbox{CustomGreen}{$\uparrow$1.91}} & \makecell[l]{73.50 \\ 74.59 \gainbox{CustomGreen}{$\uparrow$1.09}} & \makecell[l]{72.68 \\ 77.60 \gainbox{CustomGreen}{$\uparrow$4.92}} \\
\cmidrule(lr){2-11}
 & AAR & \makecell{$\times$ \\ $\checkmark$} & \makecell[l]{73.73 \\ \textemdash} & \makecell[l]{66.10 \\ 69.49 \gainbox{CustomGreen}{$\uparrow$3.39}} & \makecell[l]{69.49 \\ 71.19 \gainbox{CustomGreen}{$\uparrow$1.70}} & \makecell[l]{70.34 \\ 76.27 \gainbox{CustomGreen}{$\uparrow$5.93}} & \makecell[l]{69.49 \\ 75.42 \gainbox{CustomGreen}{$\uparrow$5.93}} & \makecell[l]{65.25 \\ 71.19 \gainbox{CustomGreen}{$\uparrow$5.94}} & \makecell[l]{67.80 \\ 69.49 \gainbox{CustomGreen}{$\uparrow$1.69}} & \makecell[l]{68.64 \\ 72.03 \gainbox{CustomGreen}{$\uparrow$3.39}} \\
\cmidrule(lr){2-11}
 & SIU & \makecell{$\times$ \\ $\checkmark$} & \makecell[l]{66.42 \\ \textemdash} & \makecell[l]{61.19 \\ 63.43 \gainbox{CustomGreen}{$\uparrow$2.24}} & \makecell[l]{61.57 \\ 66.42 \gainbox{CustomGreen}{$\uparrow$4.85}} & \makecell[l]{57.09 \\ 57.09 \gainbox{yellow!30}{$\uparrow$0.00}} & \makecell[l]{61.19 \\ 65.67 \gainbox{CustomGreen}{$\uparrow$4.48}} & \makecell[l]{61.57 \\ 66.79 \gainbox{CustomGreen}{$\uparrow$5.22}} & \makecell[l]{58.21 \\ 61.57 \gainbox{CustomGreen}{$\uparrow$3.36}} & \makecell[l]{60.45 \\ 64.55 \gainbox{CustomGreen}{$\uparrow$4.10}} \\
\cmidrule(lr){2-11}
 & VTI & \makecell{$\times$ \\ $\checkmark$} & \makecell[l]{64.79 \\ \textemdash} & \makecell[l]{61.27 \\ 63.38 \gainbox{CustomGreen}{$\uparrow$2.11}} & \makecell[l]{62.68 \\ 66.20 \gainbox{CustomGreen}{$\uparrow$3.52}} & \makecell[l]{63.03 \\ 65.49 \gainbox{CustomGreen}{$\uparrow$2.46}} & \makecell[l]{56.34 \\ 59.15 \gainbox{CustomGreen}{$\uparrow$2.81}} & \makecell[l]{57.39 \\ 58.80 \gainbox{CustomGreen}{$\uparrow$1.41}} & \makecell[l]{58.45 \\ 62.32 \gainbox{CustomGreen}{$\uparrow$3.87}} & \makecell[l]{62.68 \\ 69.01 \gainbox{CustomGreen}{$\uparrow$6.33}} \\
\midrule
\multirow{4}{=}{InternVL3.5-14B-Instruct} & PR & \makecell{$\times$ \\ $\checkmark$} & \makecell[l]{71.31 \\ \textemdash} & \makecell[l]{65.57 \\ 66.39 \gainbox{CustomGreen}{$\uparrow$0.82}} & \makecell[l]{66.67 \\ 68.03 \gainbox{CustomGreen}{$\uparrow$1.36}} & \makecell[l]{65.85 \\ 66.94 \gainbox{CustomGreen}{$\uparrow$1.09}} & \makecell[l]{58.74 \\ 61.20 \gainbox{CustomGreen}{$\uparrow$2.46}} & \makecell[l]{62.57 \\ 63.93 \gainbox{CustomGreen}{$\uparrow$1.36}} & \makecell[l]{56.01 \\ 58.20 \gainbox{CustomGreen}{$\uparrow$2.19}} & \makecell[l]{66.39 \\ 67.49 \gainbox{CustomGreen}{$\uparrow$1.10}} \\
\cmidrule(lr){2-11}
 & AAR & \makecell{$\times$ \\ $\checkmark$} & \makecell[l]{66.10 \\ \textemdash} & \makecell[l]{59.32 \\ 61.86 \gainbox{CustomGreen}{$\uparrow$2.54}} & \makecell[l]{61.02 \\ 62.71 \gainbox{CustomGreen}{$\uparrow$1.69}} & \makecell[l]{52.54 \\ 55.93 \gainbox{CustomGreen}{$\uparrow$3.39}} & \makecell[l]{61.86 \\ 63.56 \gainbox{CustomGreen}{$\uparrow$1.70}} & \makecell[l]{53.39 \\ 56.78 \gainbox{CustomGreen}{$\uparrow$3.39}} & \makecell[l]{56.78 \\ 60.17 \gainbox{CustomGreen}{$\uparrow$3.39}} & \makecell[l]{61.86 \\ 62.71 \gainbox{CustomGreen}{$\uparrow$0.85}} \\
\cmidrule(lr){2-11}
 & SIU & \makecell{$\times$ \\ $\checkmark$} & \makecell[l]{54.85 \\ \textemdash} & \makecell[l]{49.63 \\ 50.75 \gainbox{CustomGreen}{$\uparrow$1.12}} & \makecell[l]{53.36 \\ 54.85 \gainbox{CustomGreen}{$\uparrow$1.49}} & \makecell[l]{53.73 \\ 54.85 \gainbox{CustomGreen}{$\uparrow$1.12}} & \makecell[l]{47.01 \\ 48.51 \gainbox{CustomGreen}{$\uparrow$1.50}} & \makecell[l]{51.49 \\ 52.61 \gainbox{CustomGreen}{$\uparrow$1.12}} & \makecell[l]{45.90 \\ 47.76 \gainbox{CustomGreen}{$\uparrow$1.86}} & \makecell[l]{50.37 \\ 52.61 \gainbox{CustomGreen}{$\uparrow$2.24}} \\
\cmidrule(lr){2-11}
 & VTI & \makecell{$\times$ \\ $\checkmark$} & \makecell[l]{56.69 \\ \textemdash} & \makecell[l]{57.75 \\ 60.21 \gainbox{CustomGreen}{$\uparrow$2.46}} & \makecell[l]{53.52 \\ 54.58 \gainbox{CustomGreen}{$\uparrow$1.06}} & \makecell[l]{50.35 \\ 52.11 \gainbox{CustomGreen}{$\uparrow$1.76}} & \makecell[l]{50.00 \\ 51.41 \gainbox{CustomGreen}{$\uparrow$1.41}} & \makecell[l]{42.96 \\ 42.96 \gainbox{yellow!30}{$\uparrow$0.00}} & \makecell[l]{40.14 \\ 42.96 \gainbox{CustomGreen}{$\uparrow$2.82}} & \makecell[l]{47.54 \\ 51.06 \gainbox{CustomGreen}{$\uparrow$3.52}} \\
\bottomrule
\end{tabular}%
}
\caption{Effectiveness of MedVL-XLRepE across models, scenarios, and target languages. Each cell reports accuracy (\%) for the baseline ($\times$) and MedVL-XLRepE ($\checkmark$).}
\label{tab:xlrepe_result}
\end{table*}

\paragraph{Medical Vectors}
To capture finer cross-lingual variation tied to medically meaningful structure, we partition the samples of each scenario along 2 medical factors that reflect the primary clinical axes along which the scenario's visual evidence and question are jointly organized. The first factor is \emph{anatomy}, a shared factor across all scenarios that defines the spatial context of the visual evidence. The second factor is scenario-specific: \emph{pathology} for PR and SIU, where the evidence is organized around a queried clinical finding (single-image recognition for PR, temporal comparison of findings for SIU); \emph{radiological attribute} for AAR, where the question binds a finding to a queried attribute such as laterality, spatial extent, or severity; and \emph{clinical context type} for VTI, where auxiliary clinical text provides a diagnostic prior over the same visual evidence.

For each scenario $s$ and medical factor $i$, we group $S_s$ according to factor $i$ into disjoint subsets $\{S_{s,i,j}\}_{j}$. For each group we compute the group-mean cross-lingual shift
\begin{equation}
\bar{\delta}_{s,i,j}^{(l)}
=
\frac{1}{|S_{s,i,j}|}
\sum_{x\in S_{s,i,j}}
\delta_{x,tgt}^{(l)}
\end{equation}
and group centroid of the target-language hidden states
\begin{equation}
\mu_{s,i,j}^{(l)}
=
\frac{1}{|S_{s,i,j}|}
\sum_{x\in S_{s,i,j}}
h_{x,tgt}^{(l)}
\end{equation}

At inference, for a target-language input $X_{tgt}$ with hidden state $h_{x,tgt}^{(l)}$, we select for each factor $i$ the group whose centroid is most similar to $h_{x,tgt}^{(l)}$ by maximizing cosine similarity:
\begin{equation}
j_{sel}
=
\arg\max_{j}
\frac{(h_{x,tgt}^{(l)})^{\top}\,\mu_{s,i,j}^{(l)}}
     {\lVert h_{x,tgt}^{(l)}\rVert_{2}\,\lVert \mu_{s,i,j}^{(l)}\rVert_{2}}
\end{equation}
The medical vector for factor $i$ is then the group-mean shift of the selected group:
\begin{equation}
v_{med,s,i}^{(l)}
=
\bar{\delta}_{s,i,j_{sel}}^{(l)}
\end{equation}


\subsection{Cross-lingual Representation Engineering}
\label{sec:REPE}

For $X_{tgt}$ in scenario $s$, the intervention vector $v_s^{(l)}$ jointly captures the general language-level shift and the scenario-specific medical correction, averaging the latter over the two factors to balance their contributions:
\begin{equation}
v_{s}^{(l)}
= v_{lang,s}^{(l)}
  + \frac{1}{2}\sum_{i=1}^{2} v_{med,s,i}^{(l)}
\end{equation}

Prior work has shown that activation-level vector additions along specific directions can effectively steer model behavior \citep{turner2023steering}. To apply this directional correction without distorting the activation magnitude, the intervention adds $v_s^{(l)}$ with strength $\lambda$ to $h_{x,tgt}^{(l)}$ and renormalizes to the original $\ell_2$ norm:
\begin{equation}
\tilde{h}_{x,tgt}^{(l)}
= \left\lVert h_{x,tgt}^{(l)}\right\rVert_2
  \cdot
  \frac{h_{x,tgt}^{(l)} + \lambda\,v_{s}^{(l)}}
       {\left\lVert h_{x,tgt}^{(l)} + \lambda\,v_{s}^{(l)}\right\rVert_2}
\end{equation}
The corrected hidden state $\tilde{h}_{x,tgt}^{(l)}$ replaces $h_{x,tgt}^{(l)}$ in the forward pass, closing the cross-lingual gap along the scenario-conditioned medical axes while leaving all other activations unchanged.

\section{Experiments}

\subsection{Experimental Setup}
\label{sec:exp_setup}

We evaluate MedVL-XLRepE on the three open-source LVLMs used in Section~\ref{sec:eval_analysis}: Gemma3-12B-IT, Qwen3.5-9B, and InternVL3.5-14B-Instruct. Within each scenario, the benchmark is split into disjoint halves for calibration and evaluation, with detailed statistics provided in Appendix~\ref{sec:data_stats}. We use intervention strength $\lambda = 0.1$, and adopt the two scenario-specific medical factors described in Section~\ref{sec:vector_construction}.

\subsection{Main Results}
\label{sec:main_results}
\paragraph{All Clinical Scenarios Benefit from MedVL-XLRepE}
Table~\ref{tab:xlrepe_result} shows that MedVL-XLRepE yields a positive mean gain in all four clinical scenarios. AAR, which exhibits the largest baseline cross-lingual span, receives the largest mean gain (2.91\%). Even SIU, the most language-stable scenario, still attains a 2.51\% improvement.

\paragraph{Consistent Gains across All Three Backbones}
Averaged across all scenarios and target languages, MedVL-XLRepE raises cross-lingual accuracy by 2.38\%. The improvement holds across all three backbones, with per-model mean gains of 2.08\% on Gemma3-12B-IT, 3.25\% on Qwen3.5-9B, and 1.81\% on InternVL3.5-14B-Instruct.

\paragraph{Lower-Resource Target Languages Benefit Most}
The lower-resource group (Bulgarian, Thai, Arabic) obtains a mean gain of 2.68\%, exceeding the 2.16\% of the mid-to-high-resource group (Chinese, Spanish, French, Japanese). Bulgarian on Qwen3.5 (VTI) records the largest improvement of 6.33\%, and the resulting accuracy of 69.01\% even surpasses the corresponding English baseline of 64.79\%.

\subsection{Cross-lingual Alignment Visualization}
\begin{figure}[t]
\centering
\includegraphics[width=0.9\linewidth]{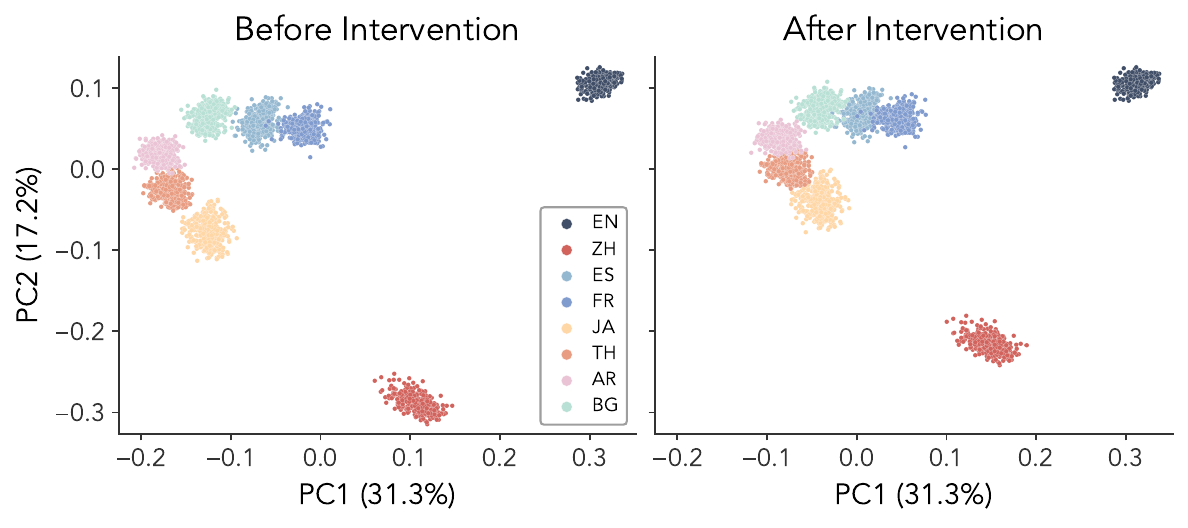}
\caption{PCA of multilingual hidden states before and after applying MedVL-XLRepE.}
\label{fig:pca}
\end{figure}

To visualize how MedVL-XLRepE reshapes the model's internal representations, we apply PCA to the final input token's hidden states at the intervention layer of Qwen3.5-9B in the PR scenario.
As shown in Figure~\ref{fig:pca}, before applying MedVL-XLRepE the eight languages form clearly separated clusters, with English isolated on one side and the seven target-language clusters lying on the
opposite side. After applying MedVL-XLRepE, every non-English cluster shifts toward the English anchor along PC1, while the English cluster itself remains unchanged by design. Averaged over the seven target languages, the centroid distance to English in PC space drops by 19.3\%.

\subsection{Hyperparameter Sensitivity Analysis}
\label{sec:layer_lambda}

\begin{figure}[t]
\centering
\includegraphics[width=\linewidth]{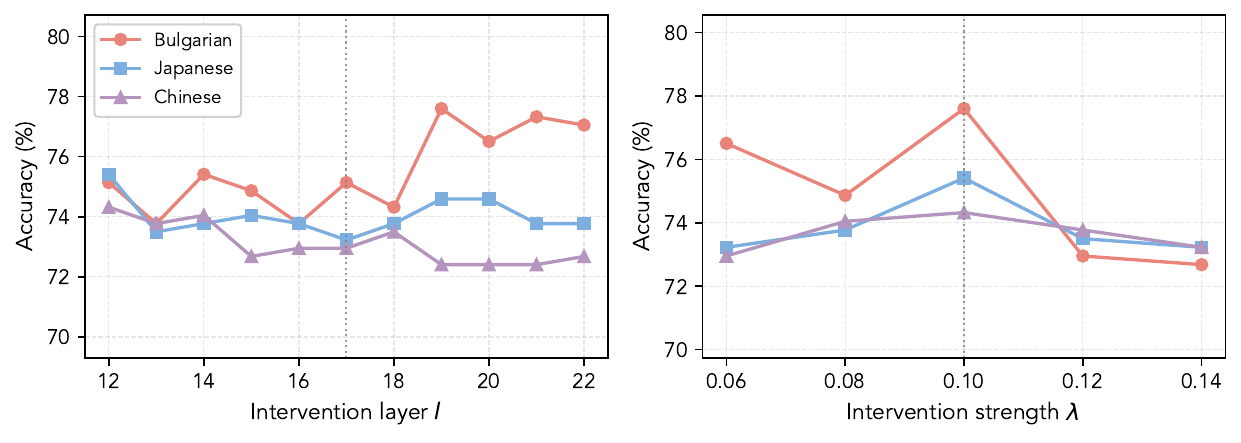}
\caption{Performance under varying intervention layer $l$ (left) and strength $\lambda$ (right) on Qwen3.5-9B/PR.}
\label{fig:layer_lambda}
\end{figure}
We analyse the sensitivity of MedVL-XLRepE to its hyper-parameters---the intervention layer $l$ and the strength $\lambda$---on Qwen3.5-9B in the PR scenario, with Bulgarian, Japanese, and Chinese as relatively low-, mid-, and high-resource target languages.

\paragraph{Intervention Layer Analysis}
Multilingual transformers concentrate language-sensitive structure and English-pivoted semantics in the middle layers~\citep{chang2022geometry,wendler2024llamas}. We therefore restrict the intervention layer $l$ to this middle-layer range. For Qwen3.5-9B, the range corresponds to layers 12--22, and Figure~\ref{fig:layer_lambda}~(left) reports the sensitivity of accuracy as $l$ is varied over these candidate layers.

\paragraph{Intervention Strength Analysis}
Activation-shift methods exhibit an inverted-U over the strength axis: too small a value fails to displace the target component, while too large a value drives the hidden state off-manifold~\citep{li2023inference}. We therefore evaluate $\lambda$ from 0.06 to 0.14 in steps of 0.02. Figure~\ref{fig:layer_lambda} (right) shows that accuracy peaks at $\lambda{=}0.10$ simultaneously for all three target languages, so we adopt $\lambda{=}0.10$ as the default value of this hyper-parameter.

\section{Conclusion}

In this work, we study cross-lingual degradation in multilingual medical VQA. To enable a fine-grained analysis, we construct a benchmark covering eight languages and organized into four representative scenarios. Evaluating five open- and closed-source LVLMs on this benchmark, we find that cross-lingual degradation is not uniform but highly scenario-dependent, and therefore cannot be addressed by a single global correction. Motivated by this finding, we propose MedVL-XLRepE, a training-free, scenario-aware cross-lingual representation engineering method that steers non-English representations toward their English counterparts at inference time. Across three LVLMs and eight languages, MedVL-XLRepE consistently mitigates cross-lingual degradation, with gains of up to 6.33\%. We hope our benchmark and method will support future research toward equitable, multilingually reliable clinical LVLMs.


\section*{Limitations}

While MedVL-XLRepE consistently mitigates cross-lingual degradation, our study has several limitations. Although our benchmark spans eight languages and four representative clinical scenarios, it still covers only a small portion of the world's languages and does not capture the full diversity of clinical tasks. In addition, MedVL-XLRepE requires white-box access to the model's internal activations, so it applies only to open-source LVLMs and cannot be used to improve closed-source LVLMs. Finally, the method depends on paired English and target-language inputs and a held-out calibration split to construct the intervention vectors, and its effectiveness when only the target language is available, or under a severe scarcity of calibration samples, is not yet characterized.

\section*{Ethical Considerations}

Our benchmark and MedVL-XLRepE are research artifacts for studying and mitigating cross-lingual degradation of LVLMs on medical VQA, and any patient-facing use would require additional clinical validation and human oversight. The benchmark is reorganized from publicly released medical VQA resources under their original licenses and access terms. Multilingual items are produced by machine translation followed by review from human experts to preserve clinical intent across languages. Equitable performance across languages is itself an ethical concern for clinical AI, and our benchmark and method are aimed at narrowing this gap so that medical LVLMs can serve patients and clinicians beyond English-speaking populations.

\bibliography{custom}

\appendix
\section{Benchmark Construction Details}

\subsection{Scenario-wise Data Statistics}
\label{sec:data_stats}
Table~\ref{tab:scenario_data_stats} reports the benchmark scale used in our main experiments. After the data sourcing, scenario-wise reorganization, multilingual translation, terminology refinement, and expert validation steps described in Section~\ref{sec:benchmark_construction}, the final benchmark contains 2,071 original English-source items and 16,568 multilingual item-language pairs under the 8-language setting. At the scenario level, PR contains 731 original items (5,848 multilingual instances), AAR contains 236 (1,888 multilingual instances), SIU contains 536 (4,288 multilingual instances), and VTI contains 568 (4,544 multilingual instances).

Within each scenario, we split the items evenly into two disjoint halves: one half is reserved as the calibration set used by MedVL-XLRepE to estimate the language-level and medical vectors in Section~\ref{sec:vector_construction}, and the other half is used as the test set on which all reported results are obtained. The split is stratified by language and by medical factor so that both halves preserve the original distribution, and no item from the calibration half ever appears in evaluation.

\begin{table}[htb]
\centering
\small
\setlength{\tabcolsep}{3pt}
\begin{tabular}{lcc}
\toprule
Scenario & Original & Multilingual \\
\midrule
Perceptual Recognition  & 731 & 5,848 \\
Attribute-Aware Recognition & 236 & 1,888 \\
Sequential Images Understanding & 536 & 4,288 \\
Vision-Text Integrated Reasoning & 568 & 4,544 \\
\midrule
Total & 2,071 & 16,568 \\
\bottomrule
\end{tabular}
\caption{Scenario-wise benchmark statistics under the 8-language setting.}
\label{tab:scenario_data_stats}
\end{table}

\subsection{Terminology Refinement Prompt}

To refine the initial Google Translate outputs described in Step 2 of Section~\ref{sec:benchmark_construction}, we use GPT-5.4 with a terminology-focused prompt. The prompt is designed to preserve the original medical intent while correcting non-standard or inaccurate medical wording in the draft translation. The full prompt template is shown in \textit{Terminology Refinement Prompt}.

\subsection{Empirical Motivation for Terminology Refinement}

To support the terminology-refinement step in Section~\ref{sec:benchmark_construction}, we quantify how often GPT-5.4 revises the initial Google Translate output in the PR and AAR scenarios. Table~\ref{tab:translation_change_rates} reports these revision rates. The results show substantial modification rates across all seven target languages used in our main experiments, indicating that raw machine translation frequently leaves terminology that is further corrected during refinement. This pattern provides empirical support for including a dedicated terminology-alignment stage in the benchmark construction pipeline.

\begin{table*}[htb]
\centering
\small
\begin{tabular}{lccccccc}
\toprule
Scenario & Chinese & Spanish & French & Japanese & Thai & Arabic & Bulgarian \\
\midrule
PR       & 53.76 & 35.84 & 55.54 & 72.64 & 81.40 & 69.49 & 50.48 \\
AAR      & 65.25 & 37.29 & 57.63 & 83.90 & 82.20 & 64.41 & 58.47 \\
\bottomrule
\end{tabular}
\caption{Percentage (\%) of items whose Google Translate draft was modified by GPT-5.4 in the VQA-RAD PR and AAR subsets.}
\label{tab:translation_change_rates}
\end{table*}

\subsection{Expert Validation Protocol}
\label{sec:expert_validation}

The expert review described in Step 3 of Section~\ref{sec:benchmark_construction} is conducted by researchers from our research group with medical and linguistics backgrounds, participating as part of the internal research collaboration. Each multilingual item is independently checked under medical and linguistic criteria, and items that fail either review are either corrected with a minimal edit or discarded from the final benchmark. The brief instructions are shown in \textit{Medical Reviewer Instruction} and \textit{Linguistic Reviewer Instruction}.

\section{Additional Analysis Results}

\subsection{Aggregated Performance Statistics}

To support the analysis of cross-lingual variation at a level more interpretable than the raw model-by-language results, we further aggregate performance by source group and by scenario. Table~\ref{tab:source_group_aggregated_language_performance} shows that closed-source models are not only stronger overall than open-source models, but also more stable across languages. Table~\ref{tab:scenario_aggregated_language_performance} shows that the cross-lingual gap is not uniform across scenarios: AAR and VTI display the largest spans, whereas PR and SIU are more stable. These aggregated results provide the direct numerical basis for the main-text comparisons between source-group robustness and scenario sensitivity.

\begin{table*}[htb]
\centering
\small
\begin{tabular}{lcccccccccc}
\toprule
Group & EN & ZH & ES & FR & JA & TH & AR & BG & Avg & Max--Min \\
\midrule
Open-source & 62.25 & 58.37 & 58.42 & 56.82 & 55.79 & 55.65 & 54.08 & 56.38 & 57.22 & 8.17 \\
Closed-source & 75.97 & 73.35 & 75.33 & 72.46 & 73.09 & 74.36 & 74.47 & 73.19 & 74.03 & 3.51 \\
\bottomrule
\end{tabular}
\caption{Aggregated language performance (\%) by source group.}
\label{tab:source_group_aggregated_language_performance}
\end{table*}

\begin{table*}[htb]
\centering
\small
\begin{tabular}{lcccccccccc}
\toprule
Setting & EN & ZH & ES & FR & JA & TH & AR & BG & Avg & Max--Min \\
\midrule
Overall & 67.73 & 64.36 & 65.19 & 63.08 & 62.71 & 63.13 & 62.23 & 63.10 & 63.94 & 5.50 \\
PR & 74.43 & 71.58 & 72.35 & 71.53 & 68.74 & 71.80 & 69.24 & 70.49 & 71.27 & 5.68 \\
AAR & 69.15 & 65.42 & 65.59 & 61.02 & 65.93 & 63.56 & 65.59 & 64.40 & 65.08 & 8.14 \\
SIU & 59.26 & 55.00 & 57.31 & 55.67 & 54.33 & 56.05 & 54.11 & 54.70 & 55.80 & 5.15 \\
VTI & 68.10 & 65.42 & 65.49 & 64.08 & 61.83 & 61.13 & 60.00 & 62.82 & 63.61 & 8.10 \\
\bottomrule
\end{tabular}
\caption{Scenario-wise aggregated language performance (\%).}
\label{tab:scenario_aggregated_language_performance}
\end{table*}

\begin{table*}[htb]
\centering
\small
\begin{tabular}{lcccccccc}
\toprule
Scenario & Rank 1 & Rank 2 & Rank 3 & Rank 4 & Rank 5 & Rank 6 & Rank 7 & Rank 8 \\
\midrule
Overall & EN (67.73) & ES (65.19) & ZH (64.36) & TH (63.13) & BG (63.10) & FR (63.08) & JA (62.71) & AR (62.23) \\
PR      & EN (74.43) & ES (72.35) & TH (71.80) & ZH (71.58) & FR (71.53) & BG (70.49) & AR (69.24) & JA (68.74) \\
AAR     & EN (69.15) & JA (65.93) & ES (65.59) & AR (65.59) & ZH (65.42) & BG (64.40) & TH (63.56) & FR (61.02) \\
SIU     & EN (59.26) & ES (57.31) & TH (56.05) & FR (55.67) & ZH (55.00) & BG (54.70) & JA (54.33) & AR (54.11) \\
VTI     & EN (68.10) & ES (65.49) & ZH (65.42) & FR (64.08) & BG (62.82) & JA (61.83) & TH (61.13) & AR (60.00) \\
\bottomrule
\end{tabular}
\caption{Scenario-wise language rankings averaged over all evaluated models.}
\label{tab:language_rankings_by_scenario}
\end{table*}

\begin{table*}[htb]
\centering
\scriptsize
\renewcommand{\arraystretch}{0.95}
\setlength{\tabcolsep}{3pt}
\resizebox{\textwidth}{!}{%
\begin{tabular}{>{\centering\arraybackslash}m{1.45cm} >{\raggedright\arraybackslash}m{2.0cm} >{\raggedright\arraybackslash}m{1.45cm} >{\raggedright\arraybackslash}m{1.45cm} >{\raggedright\arraybackslash}m{1.45cm} >{\raggedright\arraybackslash}m{1.45cm} >{\raggedright\arraybackslash}m{1.45cm} >{\raggedright\arraybackslash}m{1.45cm} >{\raggedright\arraybackslash}m{1.45cm}}
\toprule
\textbf{Scenario} & \textbf{REPE} & \textbf{ZH} & \textbf{ES} & \textbf{FR} & \textbf{JA} & \textbf{TH} & \textbf{AR} & \textbf{BG} \\
\midrule
PR
  & \makecell[l]{Baseline \\ Language \\ Medical \\ MedVL-XLRepE}
  & \makecell[l]{72.95 \\ 73.77 \gainbox{CustomGreen}{$\uparrow$0.82} \\ 73.50 \gainbox{CustomGreen}{$\uparrow$0.55} \\ 74.32 \gainbox{CustomGreen}{$\uparrow$1.37}}
  & \makecell[l]{75.96 \\ 76.23 \gainbox{CustomGreen}{$\uparrow$0.27} \\ 75.41 \gainbox{red!20}{$\downarrow$0.55} \\ 77.60 \gainbox{CustomGreen}{$\uparrow$1.64}}
  & \makecell[l]{74.86 \\ 77.05 \gainbox{CustomGreen}{$\uparrow$2.19} \\ 78.69 \gainbox{CustomGreen}{$\uparrow$3.83} \\ 76.23 \gainbox{CustomGreen}{$\uparrow$1.37}}
  & \makecell[l]{71.31 \\ 73.22 \gainbox{CustomGreen}{$\uparrow$1.91} \\ 73.22 \gainbox{CustomGreen}{$\uparrow$1.91} \\ 75.41 \gainbox{CustomGreen}{$\uparrow$4.10}}
  & \makecell[l]{75.41 \\ 75.41 \gainbox{yellow!30}{$\uparrow$0.00} \\ 74.59 \gainbox{red!20}{$\downarrow$0.82} \\ 77.32 \gainbox{CustomGreen}{$\uparrow$1.91}}
  & \makecell[l]{73.50 \\ 73.22 \gainbox{red!20}{$\downarrow$0.28} \\ 74.59 \gainbox{CustomGreen}{$\uparrow$1.09} \\ 74.59 \gainbox{CustomGreen}{$\uparrow$1.09}}
  & \makecell[l]{72.68 \\ 73.77 \gainbox{CustomGreen}{$\uparrow$1.09} \\ 75.41 \gainbox{CustomGreen}{$\uparrow$2.73} \\ 77.60 \gainbox{CustomGreen}{$\uparrow$4.92}} \\
\midrule
AAR
  & \makecell[l]{Baseline \\ Language \\ Medical \\ MedVL-XLRepE}
  & \makecell[l]{66.10 \\ 67.80 \gainbox{CustomGreen}{$\uparrow$1.70} \\ 66.95 \gainbox{CustomGreen}{$\uparrow$0.85} \\ 69.49 \gainbox{CustomGreen}{$\uparrow$3.39}}
  & \makecell[l]{69.49 \\ 69.49 \gainbox{yellow!30}{$\uparrow$0.00} \\ 68.64 \gainbox{red!20}{$\downarrow$0.85} \\ 71.19 \gainbox{CustomGreen}{$\uparrow$1.70}}
  & \makecell[l]{70.34 \\ 72.03 \gainbox{CustomGreen}{$\uparrow$1.69} \\ 72.88 \gainbox{CustomGreen}{$\uparrow$2.54} \\ 76.27 \gainbox{CustomGreen}{$\uparrow$5.93}}
  & \makecell[l]{69.49 \\ 70.34 \gainbox{CustomGreen}{$\uparrow$0.85} \\ 67.80 \gainbox{red!20}{$\downarrow$1.69} \\ 75.42 \gainbox{CustomGreen}{$\uparrow$5.93}}
  & \makecell[l]{65.25 \\ 66.10 \gainbox{CustomGreen}{$\uparrow$0.85} \\ 66.95 \gainbox{CustomGreen}{$\uparrow$1.70} \\ 71.19 \gainbox{CustomGreen}{$\uparrow$5.94}}
  & \makecell[l]{67.80 \\ 66.95 \gainbox{red!20}{$\downarrow$0.85} \\ 68.64 \gainbox{CustomGreen}{$\uparrow$0.84} \\ 69.49 \gainbox{CustomGreen}{$\uparrow$1.69}}
  & \makecell[l]{68.64 \\ 68.64 \gainbox{yellow!30}{$\uparrow$0.00} \\ 68.64 \gainbox{yellow!30}{$\uparrow$0.00} \\ 72.03 \gainbox{CustomGreen}{$\uparrow$3.39}} \\
\bottomrule
\end{tabular}%
}
\caption{Component ablation of MedVL-XLRepE on Qwen3.5-9B. Each cell lists, from top to bottom: Baseline / Language only / Medical only / MedVL-XLRepE. Colored markers show change relative to Baseline.}
\label{tab:ablation_components}
\end{table*}

\begin{table}[htb]
\centering
\small
\setlength{\tabcolsep}{3.5pt}
\renewcommand{\arraystretch}{1.15}
\resizebox{\columnwidth}{!}{%
\begin{tabular}{lccccc}
\toprule
\textbf{Group} & \textbf{$n$} & \textbf{Mean (95\% CI)} & \textbf{Paired $t$} & \textbf{Wilcoxon} & \textbf{Sign} \\
\midrule
Overall            & 84 & $+2.38$ \scriptsize{[2.05, 2.71]} & $14.37^{\dagger}$ & $7.8\!\times\!10^{-15}$ & 80\,/\,0 \\
\midrule
Gemma3-12B-IT      & 28 & $+2.08$ \scriptsize{[1.53, 2.63]} & $7.73^{\dagger}$  & $8.3\!\times\!10^{-6}$  & 26\,/\,0 \\
Qwen3.5-9B         & 28 & $+3.25$ \scriptsize{[2.58, 3.93]} & $9.89^{\dagger}$  & $5.6\!\times\!10^{-6}$  & 27\,/\,0 \\
InternVL3.5-14B    & 28 & $+1.81$ \scriptsize{[1.46, 2.17]} & $10.61^{\dagger}$ & $5.6\!\times\!10^{-6}$  & 27\,/\,0 \\
\bottomrule
\end{tabular}%
}
\caption{Statistical significance of MedVL-XLRepE cross-lingual gains, computed over the paired per-configuration gains in Table~\ref{tab:xlrepe_result}. Mean gain is in accuracy points (\%) with a $95\%$ confidence interval. Wilcoxon reports the two-sided signed-rank $p$-value; Sign reports the number of configurations with a positive\,/\,negative gain. $^{\dagger}$\,denotes $p<10^{-5}$ for the paired $t$-test ($p<10^{-15}$ for the Overall row).}
\label{tab:significance}
\end{table}

\subsection{Language Rankings by Scenario}

To examine whether language resource level is sufficient to predict multilingual medical VQA performance, we also report explicit language rankings for the overall benchmark and for each scenario separately. Table~\ref{tab:language_rankings_by_scenario} shows that, although English remains strongest overall, the relative ordering among the remaining languages is not fixed once the scenario is specified. This table therefore complements the aggregate averages by making visible where scenario-specific rankings no longer align with a simple resource-based expectation.

\section{MedVL-XLRepE Implementation Details}
\label{sec:medvl_xlrepe_impl}

\subsection{Medical Factor Extraction}
\label{sec:factor_extraction}

MedVL-XLRepE partitions calibration samples by scenario-specific medical factors (Section~\ref{sec:vector_construction}). Factor labels are obtained by combining benchmark metadata fields, rule-based extraction from question text, and LLM-based classification where metadata is absent. These labels are required only on the calibration split to form the groups $\{S_{s,i,j}\}_{j}$; at test time the group is selected purely by cosine similarity between the input hidden state and the precomputed group centroids, so no ground-truth medical labels are needed on the evaluation samples.

\textbf{Anatomy} is shared across all scenarios. Labels are derived from benchmark metadata where available (e.g., \textit{chest}, \textit{head}, \textit{abdomen}) and supplemented by rule-based extraction from question text for scenarios where explicit organ metadata is absent.

\textbf{Pathology} (PR, SIU) groups samples by the specific clinical finding being queried. For PR, samples are partitioned by the finding type targeted in the question — e.g., pneumothorax, cardiomegaly, or atelectasis — since different findings induce different visual targets and may be expressed with language-specific clinical terminology. For SIU, the same factor is applied within a temporal comparison context, where the finding type reflects what is being tracked across serial studies.

\textbf{Radiological attribute} (AAR) groups samples by the type of visual property the question asks the model to assess. Representative sub-types include laterality (e.g., left vs.\ right positioning), spatial extent (e.g., mass size or inspiratory effort), morphological attributes (e.g., contour symmetry), and density or signal intensity (e.g., hyperattenuation).

\textbf{Clinical context type} (VTI) groups samples by the type of auxiliary clinical text accompanying the image. Representative text types include structured patient case descriptions with clinical history, ECG interpretation contexts, imaging-based diagnostic scenarios, and procedural case descriptions, each providing a different form of diagnostic prior over the visual evidence.

\subsection{Component Ablation: Language Vector vs.\ Medical Vector}
\label{sec:ablation_components}

MedVL-XLRepE applies two corrections: a language-level vector that steers target-language representations toward English, and a medical vector that re-centres them within the scenario-specific factor subspace. Table~\ref{tab:ablation_components} isolates each component on Qwen3.5-9B in the PR and AAR scenarios. Applied alone, each component yields limited average gains and occasionally hurts individual languages. For instance, the medical vector alone decreases Japanese by 1.69\% in AAR, and the language vector alone decreases Arabic by 0.28\% in PR. MedVL-XLRepE eliminates these isolated regressions and achieves +2.34\% on PR and +3.99\% on AAR. The AAR gain notably exceeds the sum of the two isolated contributions (+1.09\%), indicating that the two corrections reinforce each other when applied jointly.

\section{Statistical Significance of MedVL-XLRepE Gains}
\label{sec:significance}

This appendix examines whether the cross-lingual improvements reported in Table~\ref{tab:xlrepe_result} are directionally consistent across the evaluated model--scenario--language configurations.

\subsection{Test Protocol}
\label{sec:significance_protocol}

All evaluations use greedy decoding (temperature $0$), and MedVL-XLRepE is a fixed inference-time intervention with no stochastic component; each (model, scenario, target-language) configuration therefore yields a single deterministic accuracy that repeated runs reproduce exactly. The relevant question is consequently not run-to-run variance, but whether the improvement is \emph{systematic} across the population of evaluated configurations.

We treat each of the $84$ configurations in Table~\ref{tab:xlrepe_result} ($3$ backbones $\times\,4$ scenarios $\times\,7$ target languages; English is the anchor and is not intervened) as a paired observation, pairing the baseline accuracy with the MedVL-XLRepE accuracy, and define the per-configuration gain as their difference. On these $84$ paired gains we apply three complementary tests: a paired $t$-test (parametric), a Wilcoxon signed-rank test (non-parametric, no normality assumption), and a sign test (distribution-free). To ensure that significance is not an artefact of pooling across heterogeneous backbones, we additionally report each test separately for every model ($28$ configurations each).

\subsection{Results}
\label{sec:significance_results}

Table~\ref{tab:significance} summarizes the analysis. Across all $84$ configurations the mean gain is $+2.38\%$ with a $95\%$ confidence interval of $[+2.05, +2.71]$ that lies well above zero, and the paired $t$-test strongly rejects the no-effect null ($t(83)=14.37$, $p<10^{-15}$). The non-parametric tests agree: the Wilcoxon signed-rank test gives $z=7.77$ ($p=7.8\times10^{-15}$), and of the $84$ configurations $80$ show a positive gain, $4$ show exactly no change, and none shows a decrease, so the sign test likewise rejects at $p<10^{-15}$.

The effect holds within every backbone. The per-model mean gains are $+2.08\%$, $+3.25\%$, and $+1.81\%$ for Gemma3-12B-IT, Qwen3.5-9B, and InternVL3.5-14B-Instruct respectively, each with a $95\%$ confidence interval strictly above zero and each individually significant under all three tests. Significance is therefore not an artefact of pooling across models. Moreover, since no configuration in the entire evaluation grid exhibits a negative gain, the improvement delivered by MedVL-XLRepE is directionally consistent across all models, scenarios, and target languages.

\begin{figure*}[t]
\centering
\begin{tcolorbox}[
    width=\textwidth,
    colframe=black, 
    fonttitle=\bfseries,
    title=\textit{Terminology Refinement Prompt}
]
\label{prompt:terminology_refinement}
You are a medical translation reviewer. You are given an English medical source text and a draft translation produced by Google Translate. Your task is to produce the final translation in the target language. \\

Requirements: \\
1) Preserve the exact medical meaning, clinical intent, and level of specificity of the English source. Do not add, omit, weaken, or reinterpret information. \\
2) Correct medical terminology using standard, clinically appropriate wording in the target language. \\
3) Keep the original discourse form. If the source is a question, the output must remain a question. If the source is an answer option or fragment, preserve the same sentence style and granularity. \\
4) Revise the Google Translate draft conservatively: keep wording that is already correct and natural, and change only expressions that are medically inaccurate, linguistically awkward, or non-standard in clinical usage. \\
5) Pay particular attention to terminology that can affect downstream medical reasoning, including disease names, anatomical structures, laterality, severity descriptors, procedures, and pathology-related expressions. \\
6) Return only the final corrected translation. Do not output explanations, notes, JSON, or markdown. \\

English source text: \{source\_text\} \\
Draft target-language translation: \{google\_translation\}
\end{tcolorbox}
\end{figure*}

\begin{figure*}[t]
\centering
\begin{tcolorbox}[
    width=\textwidth,
    colframe=black,
    fonttitle=\bfseries,
    title=\textit{Medical Reviewer Instruction}
]
\label{prompt:medical_review}
You are given an English source item from a medical VQA benchmark and its target-language version produced by machine translation and terminology refinement. Your task is to decide whether the target-language version is medically faithful to the English source. \\

For each item, verify the following: \\
1) \textbf{Medical meaning.} The target-language question and answer options preserve the exact medical meaning, clinical intent, and level of specificity of the English source. No clinical information is added, omitted, weakened, or reinterpreted. \\
2) \textbf{Terminology.} Medical terms (disease names, anatomical structures, laterality, severity descriptors, procedures, and pathology-related expressions) are standard and clinically appropriate in the target language. \\
3) \textbf{Image grounding.} The question remains answerable from the same visual evidence as the English source. The translation does not introduce visual cues, anatomical references, or attributes that are not present in the image. \\
4) \textbf{Answer correctness.} The identity of the correct answer remains the same as in the English source. \\
\end{tcolorbox}
\end{figure*}

\begin{figure*}[t]
\centering
\begin{tcolorbox}[
    width=\textwidth,
    colframe=black,
    fonttitle=\bfseries,
    title=\textit{Linguistic Reviewer Instruction}
]
\label{prompt:linguistic_review}
You are given an English source item from a medical VQA benchmark and its target-language version produced by machine translation and terminology refinement. Your task is to decide whether the target-language version is linguistically natural and structurally consistent with the source. \\

For each item, verify the following: \\
1) \textbf{Grammatical correctness.} The target-language question and answer options are grammatically well-formed in the target language. \\
2) \textbf{Naturalness.} The phrasing reads naturally to a native speaker. Awkward or literally translated constructions are revised into natural target-language expressions, without changing the meaning. \\
3) \textbf{Discourse form.} If the English source is a question, the target-language version remains a question; answer options remain options of the same sentence style and granularity. \\
4) \textbf{Answer-mapping stability.} Option order, label letters, and the structural mapping between the question and its options are preserved exactly as in the English source. \\
\end{tcolorbox}
\end{figure*}

\clearpage
\begin{figure*}[!t]
\centering
\small
\begin{tcolorbox}[
    width=\textwidth,
    colframe=black,
    fonttitle=\bfseries,
    boxsep=2pt,
    left=3pt, right=3pt, top=3pt, bottom=3pt,
    title=\textit{Perceptual Recognition (PR)}
]
\begin{minipage}[c]{0.18\textwidth}
\centering
\includegraphics[height=0.11\textheight,keepaspectratio]{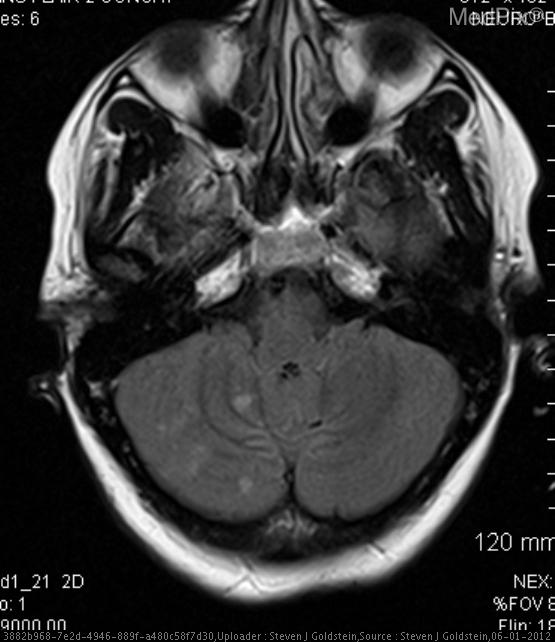}
\end{minipage}\hfill
\begin{minipage}[c]{0.78\textwidth}
\textbf{Question:} Is the Right vertebral artery normal? \\
\textbf{Answer:} No
\end{minipage}
\end{tcolorbox}

\begin{tcolorbox}[
    width=\textwidth,
    colframe=black,
    fonttitle=\bfseries,
    boxsep=2pt,
    left=3pt, right=3pt, top=3pt, bottom=3pt,
    title=\textit{Attribute-Aware Recognition (AAR)}
]
\begin{minipage}[c]{0.27\textwidth}
\centering
\includegraphics[width=0.92\linewidth,keepaspectratio]{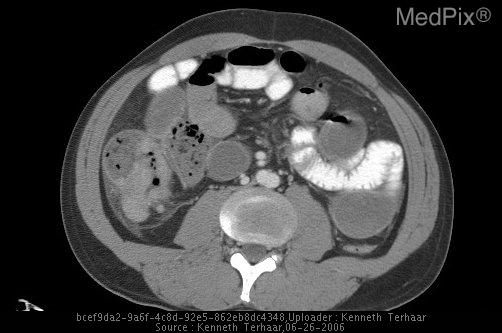}
\end{minipage}\hfill
\begin{minipage}[c]{0.69\textwidth}
\textbf{Question:} Is the mass surrounding the aorta? \\
\textbf{Answer:} No
\end{minipage}
\end{tcolorbox}

\begin{tcolorbox}[
    width=\textwidth,
    colframe=black,
    fonttitle=\bfseries,
    boxsep=2pt,
    left=3pt, right=3pt, top=3pt, bottom=3pt,
    title=\textit{Sequential Images Understanding (SIU)}
]
\begin{minipage}[c]{0.30\textwidth}
\centering
\begin{minipage}[t]{0.48\linewidth}
\centering
\includegraphics[width=\linewidth,keepaspectratio]{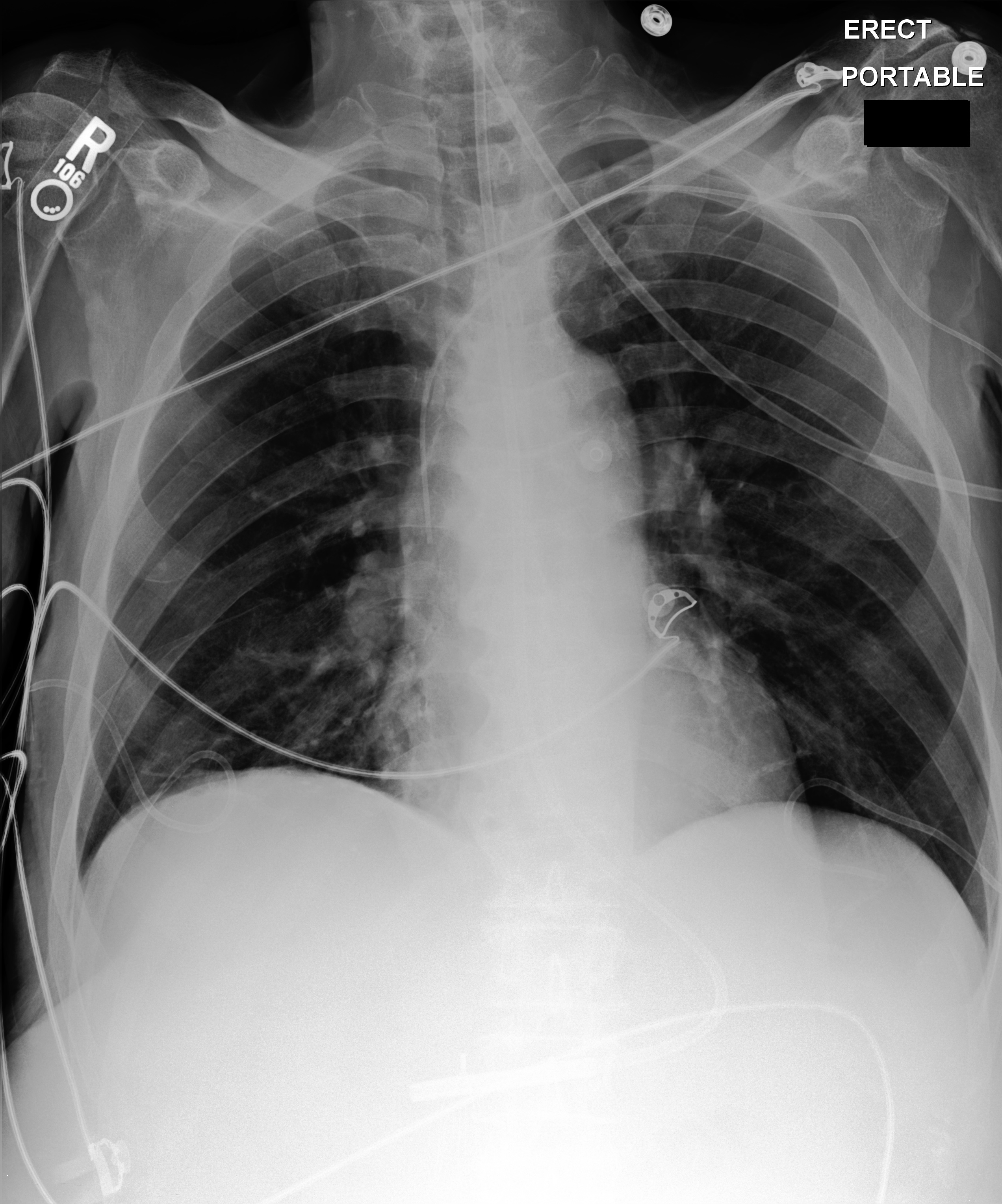}\\[-2pt]
{\scriptsize Prior}
\end{minipage}\hfill
\begin{minipage}[t]{0.48\linewidth}
\centering
\includegraphics[width=\linewidth,keepaspectratio]{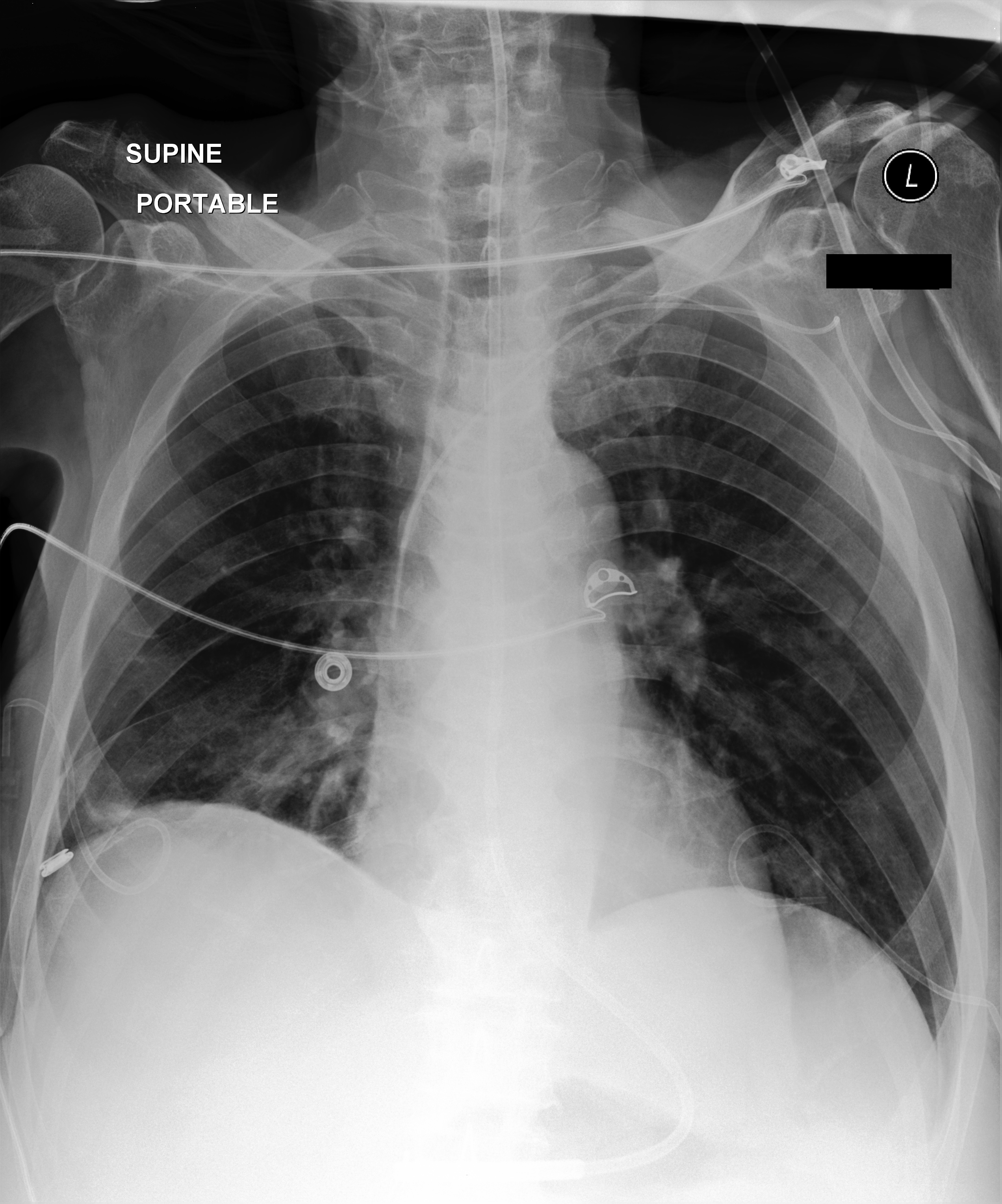}\\[-2pt]
{\scriptsize Current}
\end{minipage}
\end{minipage}\hfill
\begin{minipage}[c]{0.66\textwidth}
\textbf{Question:} How has the aeration changed in the base of the right lung when comparing both CXR images? \\
\textbf{Options:} \\
A. There is worsened aeration in the right lung base. \\
B. There is no change in aeration in the right lung base. \\
C. There is improved aeration in the right lung base. \\
D. Aeration has completely improved in both lung bases. \\
\textbf{Answer:} C. There is improved aeration in the right lung base.
\end{minipage}
\end{tcolorbox}

\begin{tcolorbox}[
    width=\textwidth,
    colframe=black,
    fonttitle=\bfseries,
    boxsep=2pt,
    left=3pt, right=3pt, top=3pt, bottom=3pt,
    title=\textit{Vision-Text Integrated Reasoning (VTI)}
]
\begin{minipage}[c]{0.23\textwidth}
\centering
\includegraphics[width=\linewidth,height=0.16\textheight,keepaspectratio]{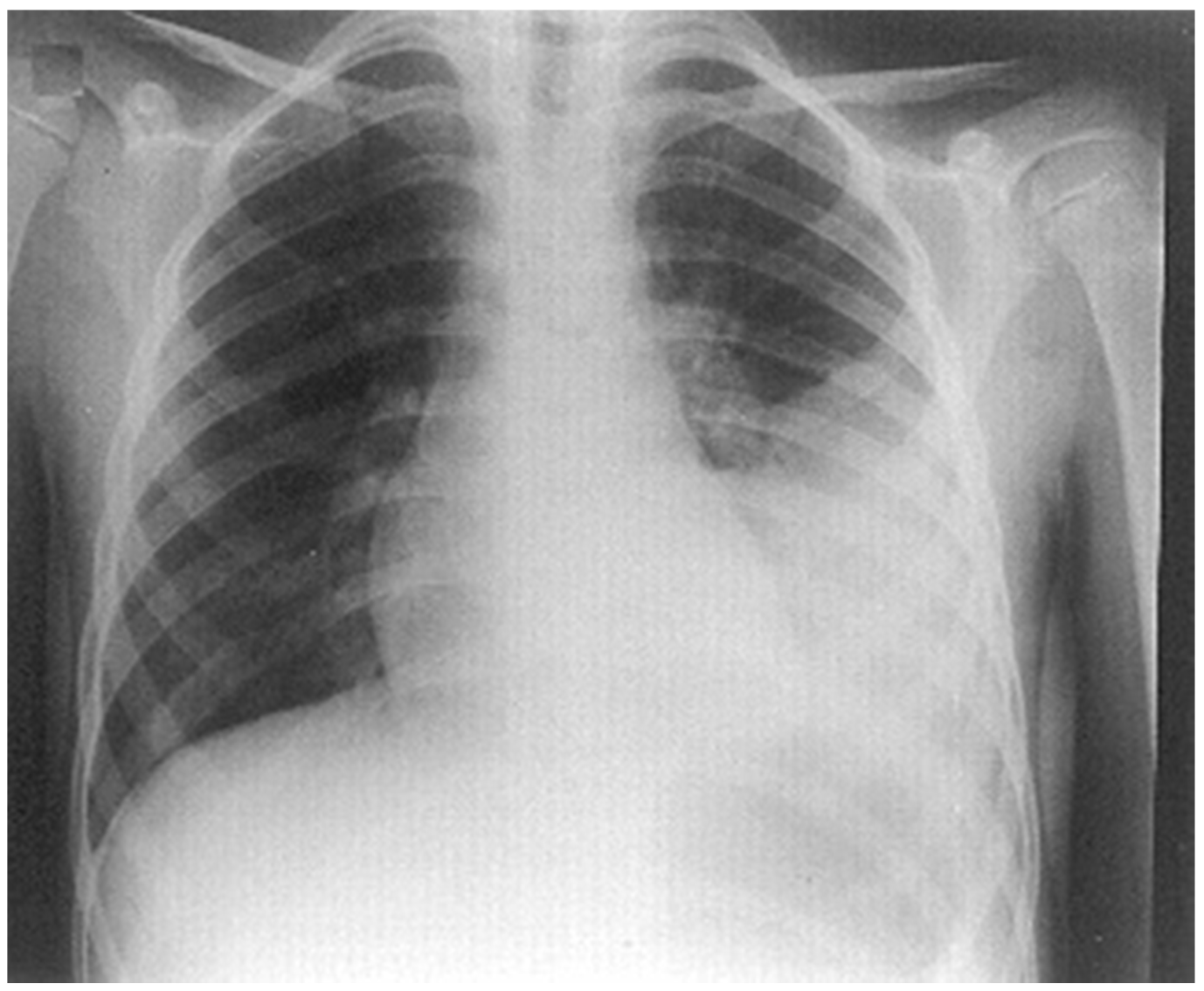}
\end{minipage}\hfill
\begin{minipage}[c]{0.73\textwidth}
\textbf{Question:} A three-year-old child, malnourished, who had been hospitalized ten days ago, is taken to the Medical Emergency Department. The child has had an unchecked fever, cough and difficulty breathing for two days. The mother reports that the patient is unable to drink liquids and has vomited several times in the last 24 hours. Upon physical examination, the doctor observed that the child had a regular general condition, fever of 38.5$^{\circ}$C, mild dehydration, tachydyspnea, with intercostal insufficiency, presence of crackling rales and decreased breath sounds in the left hemithorax; heart rate = 130 bpm, respiratory rate = 64 bpm and oxygen saturation = 91\%. The chest x-ray is shown below. The etiological agent and treatment of pneumonia presented by the child are: \\
\textbf{Options:} \\
A. \textit{Haemophilus influenzae}; crystalline penicillin. \\
B. \textit{Streptococcus pneumoniae}; procaine penicillin. \\
C. \textit{Staphylococcus aureus}; ceftriaxone associated with oxacillin. \\
D. \textit{Mycoplasma pneumoniae}; antibiotic therapy with macrolides. \\
\textbf{Answer:} C. \textit{Staphylococcus aureus}; ceftriaxone associated with oxacillin.
\end{minipage}
\end{tcolorbox}
\caption{Examples of the four clinical scenarios.}
\label{fig:scenario_examples}
\end{figure*}

\end{document}